\documentclass[conference]{IEEEtran}
\IEEEoverridecommandlockouts

\usepackage{cite}
\usepackage{amsmath,amssymb,amsfonts}
\usepackage{algorithmic}
\usepackage{graphicx}
\usepackage{textcomp}
\usepackage{xcolor}
\usepackage{spconf,amsmath,graphicx}
\usepackage{multirow}
\usepackage{tabularray}
\usepackage{subfig}
\usepackage{algorithm}
\usepackage{algorithmic}
\usepackage{amssymb}
\usepackage{booktabs}
\usepackage{colortbl}
\usepackage{pifont}       
\usepackage{bbding}       
\usepackage{fontawesome}  
\def\BibTeX{{\rm B\kern-.05em{\sc i\kern-.025em b}\kern-.08em
    T\kern-.1667em\lower.7ex\hbox{E}\kern-.125emX}}
\usepackage{stfloats}
\renewcommand{\checkmark}{\ding{51}}%

\begin{document}

\title{Diversified Augmentation with Domain Adaptation for Debiased Video Temporal Grounding\\
\thanks{* Corresponding author.

This research is supported by the National Natural Science Foundation of China (No. 62406267), Guangzhou-HKUST(GZ) Joint Funding Program (Grant No.2023A03J0008), Education Bureau of Guangzhou Municipality and the Guangzhou Municipal Education Project (No. 2024312122).}
}

\author{
\IEEEauthorblockN{Junlong Ren\textsuperscript{1}, Gangjian Zhang\textsuperscript{1}, Haifeng Sun\textsuperscript{2}, Hao Wang\textsuperscript{1*}}
\IEEEauthorblockA{\textsuperscript{1}\textit{The Hong Kong University of Science and Technology (Guangzhou)}, Guangzhou, China \\
\textsuperscript{2}\textit{Beijing University of Posts and Telecommunications}, Beijing, China \\
Email: \{jren686, gzhang292\}@connect.hkust-gz.edu.cn, hfsun@bupt.edu.cn, haowang@hkust-gz.edu.cn}
}

\maketitle

\begin{abstract}
Temporal sentence grounding in videos (TSGV) faces challenges due to public TSGV datasets containing significant temporal biases, which are attributed to the uneven temporal distributions of target moments.
Existing methods generate augmented videos, where target moments are forced to have varying temporal locations. However, since the video lengths of the given datasets have small variations, only changing the temporal locations results in poor generalization ability in videos with varying lengths. 
In this paper, we propose a novel training framework complemented by diversified data augmentation and a domain discriminator. The data augmentation generates videos with various lengths and target moment locations to diversify temporal distributions. However, augmented videos inevitably exhibit distinct feature distributions which may introduce noise. To address this, we design a domain adaptation auxiliary task to diminish feature discrepancies between original and augmented videos. We also encourage the model to produce distinct predictions for videos with the same text queries but different moment locations to promote debiased training. 
Experiments on Charades-CD and ActivityNet-CD datasets demonstrate the effectiveness and generalization abilities of our method in multiple grounding structures, achieving state-of-the-art results.
\end{abstract}


\section{Introduction}
\label{sec:intro}

Temporal sentence grounding in videos (TSGV) \cite{yuan2019semantic,ghosh2019excl,zeng2020dense,mun2020local,hu2021coarse,wang2021structured,zhang2020span,wang2022negative,wang2023scene,lin2023univtg,jiang2024prior,fang2024not} aims to identify a video segment most closely aligned with a specified text query within an untrimmed video. Recent studies \cite{otani2020challengesmr,lan2022closer} have indicated that temporal biases in
public TSGV datasets provide strong shortcuts for models to overfit rather than establishing the essential multi-modal alignment. 

\begin{figure}[t]
\centering
\setlength{\abovecaptionskip}{0pt} 
\setlength{\intextsep}{0pt} 
\setlength{\textfloatsep}{0pt} 
\vspace{-0.85em}
  \subfloat[Original Distribution]{\includegraphics[width=0.4\linewidth]{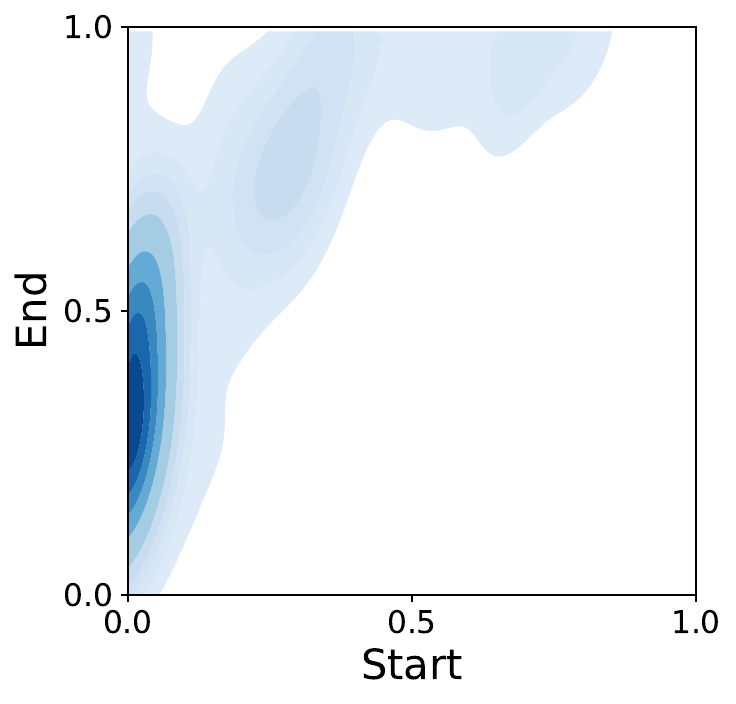}} 
  \subfloat[Augmented Distribution]{\includegraphics[width=0.52\linewidth]{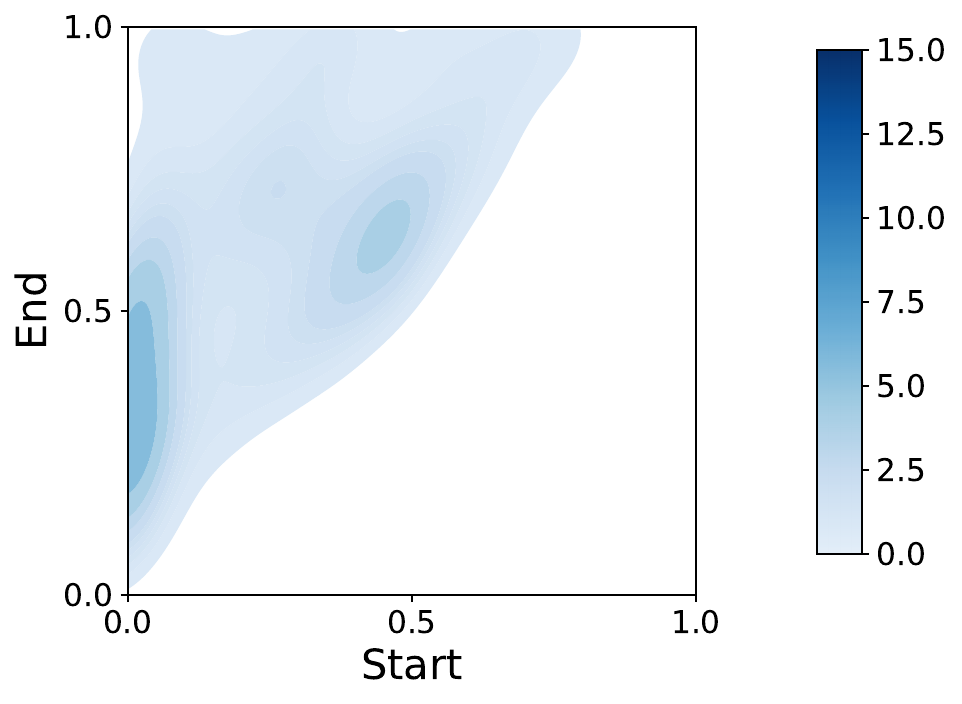}} 
\caption{Temporal distributions of target moments with the action \emph{lead} in ActivityNet Captions before and after adding our data augmentation. 
}
\label{temporal_distribution_close}
\vspace{-1.5em}
\end{figure}

Temporal biases arise from uneven temporal distributions of queries and their corresponding target moments. As shown in Fig. \ref{temporal_distribution_close} (a), in ActivityNet Captions \cite{krishna2017dense}, queries with the term \emph{lead} are primarily associated with target moments located in the first half of videos.
Consequently, the presence of \emph{lead} within a query significantly increases the probability of the target moment being predicted in the first half of the video, despite the actual target moment residing elsewhere.

Numerous works \cite{lan2022closer,hao2022can,lan2023curriculum,qi2024bias,liu2022reducing, wang2023mixup, zhang2021towards, qi2024collaborative} have been proposed to tackle the issue of temporal bias. In particular, \cite{hao2022can,lan2023curriculum} generates augmented videos through video shuffling. However, they only mitigate a limited extent of temporal biases. They primarily focus on the impact of biased target moment locations, neglecting the influence of biased video lengths. Given that the video durations within the public datasets exhibit limited fluctuations, merely altering the temporal locations leads to poor generalization ability in videos of diverse lengths. 
Besides, they often disrupt temporal continuity and logical coherence in the original video sequences, leading to model confusion and introducing noise.
Moreover, \cite{hao2022can,lan2023curriculum, zhang2021towards,qi2024bias,liu2022reducing,qi2024collaborative} only apply their methods to a certain grounding structure, limiting their contributions to diverse grounding structures.

In this paper, we propose a novel debiased training framework with diversified data augmentation and a domain adaptation auxiliary task.
The core idea of our data augmentation is to generate videos with diverse lengths and target moment locations while preserving temporal continuity and logical coherence. As depicted in Fig. \ref{temporal_distribution_close} (b), the biased temporal distribution is effectively diversified after adopting our data augmentation. 
Nevertheless, augmented videos inevitably exhibit distinct feature distributions that inadvertently introduce noise into training.
To eliminate the noise and enhance grounding precision, we design the domain adaptation task to alleviate the feature discrepancies between the original and augmented videos. 
We also combine original and augmented videos as paired input. The model is forced to make distinct predictions for these paired inputs. Although both original and augmented videos share the same text queries, they differ in the temporal locations of target moments. By doing so, the model distinguishes the temporal discrepancies between these videos, enhancing its ability to discern temporal relationships.

For clarification, we employ a span-based grounding backbone and our framework can be easily integrated into other grounding structures. Extensive experiments on Charades-CD \cite{lan2022closer} and ActivityNet-CD \cite{lan2022closer} verify the debias efficacy of our proposed framework on multiple grounding structures.

\section{Proposed Framework}
\label{sec:Proposed Framework}
\begin{figure}[t] 
		\centering 
		\includegraphics[width=1\linewidth]{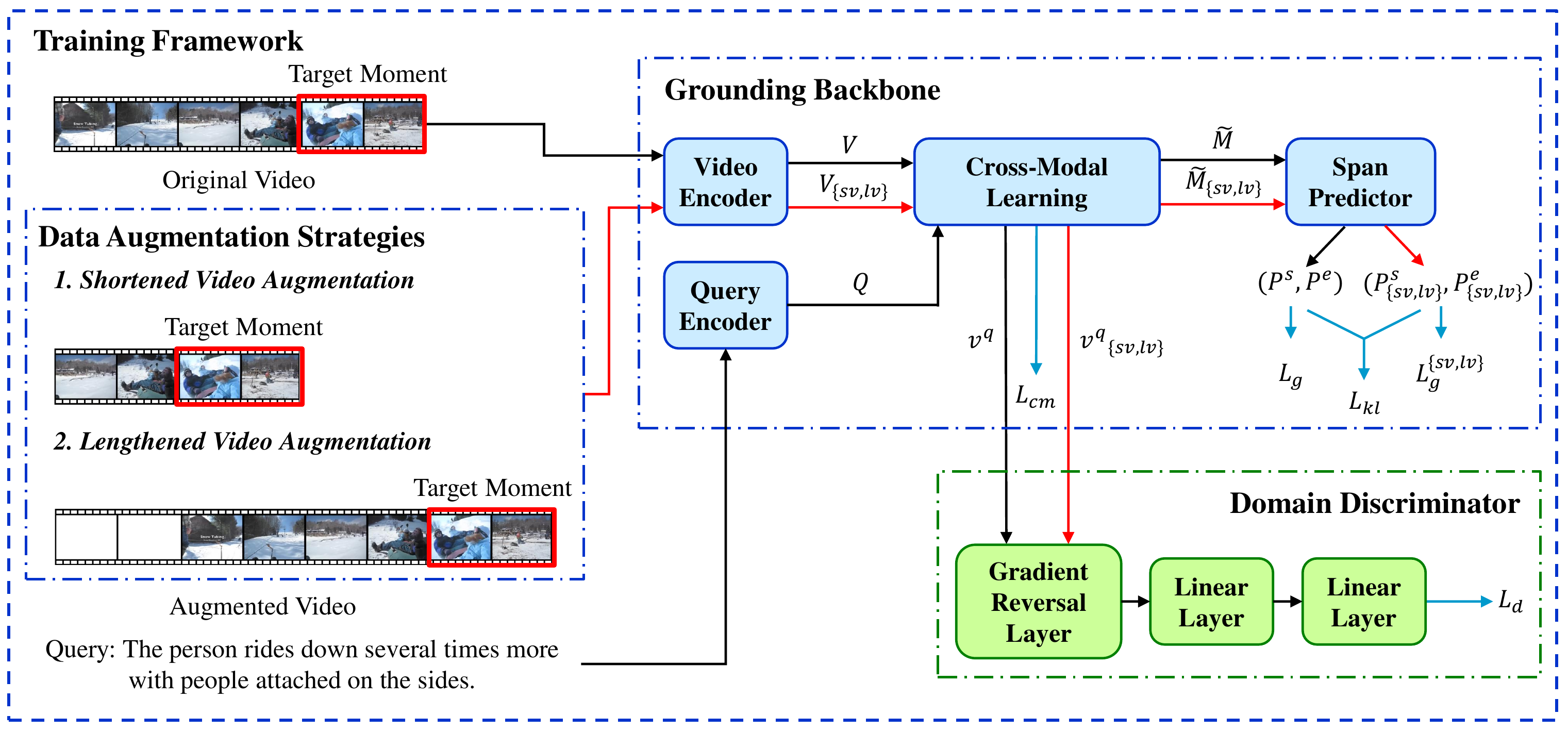} 
		\caption{The overview of our training framework.} 
		\label{overview} 
        \vspace{-1.5em}
	\end{figure}

The overview of our framework is depicted in Fig. \ref{overview}. We first delineate the formulation of TSGV. Then we expound the grounding backbone. 
To clarify, we use a span-based model with a standard transformer encoder-decoder architecture \cite{vaswani2017attention} that directly predicts index tokens in an auto-regressive manner. Our method can be adapted to other grounding structures with minor modifications.
Subsequently, we detail two novel data augmentation strategies followed by the domain adaptation module and overall training objectives.

\subsection{Problem Formulation}
Given a video ${F}_{V}$  and a text query ${F}_{S}$, a TSGV model will predict a pair of start and end timestamps $\left( {\tau^{s},\tau^{e}} \right)$ of the video segment which is semantically relevant to the text query.

\subsection{Grounding Backbone}

\paragraph{Video Encoder}
We first adopt a pre-trained I3D \cite{carreira2017quo} network to extract clip-level features, and then apply a multi-layered perceptron
(MLP) to project features into a high-level semantic space of video-language modalities. The encoded features are denoted as $V = \left\{ v_{t} \right\}_{t = 1}^{T} \in \mathbb{R}^{T \times D}$, where $T$ is the number of video clips and $D$ is the feature dimension.

\paragraph{Query Encoder}
The word-level embeddings are obtained using Glove \cite{pennington2014glove}. Another MLP is added to project the embeddings into the same high-level semantic space. We then utilize a transformer encoder \cite{vaswani2017attention} to fuse the sequential information among the word embeddings and compute the sentence-level representations, which are denoted as $Q = \left\{ q_{n} \right\}_{n = 1}^{N} \in \mathbb{R}^{N \times D}$ and $s \in \mathbb{R}^{D}$. $N$ is the number of words.

\paragraph{Cross-Modal Learning}
We first fuse $V$ and $Q$ into multi-modal features $M \in \mathbb{R}^{T \times D}$ through the co-attention mechanism \cite{yu2018qanet} and a standard transformer encoder \cite{vaswani2017attention}. 
The aggregated representation $m^{q} \in \mathbb{R}^{D}$ is computed using the [CLS] token \cite{devlin-etal-2019-bert}.
Then we predict the cross-modal relevance score to the query for each video clip. We first concatenate the sentence-level representation $s$ with each feature in $M$, which is represented as $M^{s}$. The cross-modal relevance scores $c^{m}$ are predicted through an MLP and $M$ is gated by these scores:
\begin{equation}
    c^{m} = Sigmoid\left( MLP\left( M^{s} \right) \right) \in \mathbb{R}^{T},~\overset{\sim}{M} = c^{m} \cdot M.
\end{equation}
We optimize the cross-modal learning module through the binary cross-entropy loss:
\begin{equation}
    L_{cm} = f_{BCE}\left( c^{m},c^{v} \right),
\end{equation}
where $c^{v}$ is a sequence of 0-1, values between $\tau^{s}$ to $\tau^{e}$ are assigned to 1 and the others are set to 0. 

\paragraph{Span Predictor}
We employ a transformer decoder \cite{vaswani2017attention} as the predictor. The predictor receives the multi-modal features as input and generates index token probabilities as outputs. The first two predicted index tokens are utilized as the start and end timestamps. The probability distributions are denoted as $p^{s}$ and $p^{e}$, respectively. We utilize the cross-entropy loss to supervise the span predictor:
\begin{equation}
    L_{g} = \frac{1}{2}\left( f_{CE}\left( p^{s},\tau^{s} \right) + f_{CE}\left( p^{e},\tau^{e} \right) \right).
\end{equation}

\subsection{Diversified Data Augmentation}
Note that temporal biases may occur because of uneven temporal distributions within datasets. As a result, the span predictor module struggles to generalize effectively on out-of-distribution samples. To resolve this issue, we propose to adopt data augmentation to generate videos with diverse lengths and target moment locations from two distinct perspectives. 

Technically, to achieve more diversified temporal distributions, we employ two approaches that enhance the dataset's temporal diversity while preserving the logical consistency of the original video sequences. Our proposed data augmentation strategies are outlined as follows.

\paragraph{Shortened Video Augmentation}
We randomly truncate segments before the target moment’s start timestamp for videos satisfying the condition that $\tau^{s} > \beta_{sv}$. Note that trimming video clips with a small length might not significantly alter the temporal distribution. Therefore, we pre-define a threshold $\beta_{sv}$ as the minimum truncation length. 
\begin{equation}
\begin{split}
\delta_{sv} &\sim U\left( \beta_{sv},\tau^{s} \right), V_{sv} = \left\{ v_{t} \right\}_{t = \delta_{sv} + 1}^{(T - \delta_{sv})},\\
~\tau_{sv}^{s} &= \tau^{s} - \delta_{sv},~\tau_{sv}^{e} = \tau^{e} - \delta_{sv},
\end{split}
\end{equation}
where $\delta_{sv}$ is the truncation length.
\paragraph{Lengthened Video Augmentation}
It involves inserting blank clips with random lengths at the start of videos. We pre-define a threshold  $\beta_{lv}$ as the minimum padding length.
\begin{equation}
\begin{split}
\left. \delta_{lv} \right. \sim U\left( \beta_{lv},\tau^{s} + \beta_{lv} \right)&, V_{lv} = \left\lbrack {\left\{ v_{z} \right\}_{z = 1}^{\delta_{lv}};\left\{ v_{t} \right\}_{t = 1}^{T}} \right\rbrack,\\
~\tau_{lv}^{s} = \tau^{s} + \delta_{lv}&,~\tau_{lv}^{e} = \tau^{e} + \delta_{lv},
\end{split}
\end{equation}
where $\delta_{lv}$ is the padding length and every element in $v_{z}$ is 0.

These augmented videos contain the entire target moments from the original videos, yet exhibit varying target moment temporal locations and video durations, resulting in diverse temporal distributions. 
Note that clipping videos may disrupt the semantic association for queries that need long-term context dependencies (e.g., ‘‘Child is running \emph{again}.’’), and padding blank clips at the start of videos may not cause such a phenomenon since blank clips contain no meaningful actions. Therefore, videos with queries containing certain terms that suggest long-term context dependencies (e.g., first, after, continue) are not clipped but only padded.
By adopting these strategies, the model shifts its focus from dataset temporal biases to extracting meaningful target action features.

\subsection{Domain Adaptation}

However, the process of data augmentation could inevitably cause changes in data distributions and lead to another kind of data bias. For instance, the incorporation of blank video clips may add unnecessary data that could introduce noise during training. The model may separately learn data biases of the original and augmented videos as these videos exhibit distinct data distributions that are easy to distinguish. As a result, the domain discrepancy between the original and augmented videos may impair model predictive capability.
 
To resolve this issue, we employ a domain discriminator which is equipped with a gradient reversal layer \cite{ganin2016domain}. We put the gradient reversal layer between the feature encoder and the domain discriminator. 
During the forward propagation phase, it maintains the integrity of the input data without any alterations. In contrast, during the backpropagation phase, it multiplies the gradient received from the subsequent layer by $-1$ to invert its sign before passing it to the preceding layer. 

In the optimization process, the gradient’s sign undergoes inversion within the feature encoder, yet persists without alteration in the domain discriminator. Consequently, the domain discriminator minimizes its loss function, whereas the feature encoder maximizes it. The domain discriminator is designed to differentiate between original and augmented videos. The feature encoder renders them indistinguishable to the domain discriminator, resulting in well-aligned feature distributions between original and augmented videos. 

This alignment improves the model’s ability to generalize and maintain robust predictive performance across both video categories. The domain classification scores are predicted with the aggregated representation $m^{q}$ as follows:
\begin{equation}
    o^{c} = Sigmoid\left( DomainDiscriminator\left( m^{q} \right) \right).
\end{equation}
We adopt the cross-entropy loss to optimize the domain discriminator:
\begin{equation}
    L_{d} = f_{CE}\left( o^{c},0 \right) + f_{CE}\left( o_{\{ sv,lv\}}^{c},1 \right).
\end{equation}
Note that both original and augmented videos correspond to the same text queries but exhibit discrepancies in temporal locations of target moments, we utilize the Kullback-Leibler divergence to enhance the discrepancy in prediction scores between them:
\begin{equation}
    L_{kl} = {1 - D}_{kl}\left( {p^{s} \parallel p_{\{ sv,lv\}}^{s}} \right)  { - D}_{kl}\left( {p^{e} \parallel p_{\{ sv,lv\}}^{e}} \right).
\end{equation}
The objective is to direct the model’s attention to the temporal disparities present in the well-aligned original and augmented videos, thereby improving the model’s temporal discernment.

\subsection{Training Objectives}
\paragraph{Baseline}
The training loss of the baseline model is:
\begin{equation}
    L_{loc} = L_{g} + \lambda_{1}L_{cm}.
\end{equation}

\paragraph{Ours}
The final training loss of our framework is:
\begin{equation}
    L = L_{loc} + \lambda_{2}L_{d} + \lambda_{3}L_{kl},
\end{equation}
where $\lambda_{\{1,2,3\}}$ are weight hyperparameters.

\begin{table*}[t]
\scriptsize      
\centering
  \caption{Comparison results on Charades-CD and ActivityNet-CD. * indicates our reproduced results.}
  \label{main_results}
\begin{tabular}{lccccccccccccc}
\toprule
\multirow{3}{*}{Method}   & \multirow{3}{*}{Venue} & \multicolumn{6}{c}{Charades-CD}                                                           & \multicolumn{6}{c}{ActivityNet-CD}                                                        \\  
\cmidrule(lr){3-8} \cmidrule(lr){9-14}
                          &                          & \multicolumn{2}{c}{dR@1,IoU=0.3}              & \multicolumn{2}{c}{dR@1,IoU=0.5}              & \multicolumn{2}{c}{dR@1,IoU=0.7}              & \multicolumn{2}{c}{dR@1,IoU=0.3}                 & \multicolumn{2}{c}{dR@1,IoU=0.5}              & \multicolumn{2}{c}{dR@1,IoU=0.7}                \\ 
\cmidrule(lr){3-4} \cmidrule(lr){5-6} \cmidrule(lr){7-8} \cmidrule(lr){9-10} \cmidrule(lr){11-12} \cmidrule(lr){13-14}
                          &                          & i.i.d              & o.o.d              & i.i.d              & o.o.d & i.i.d              & o.o.d & i.i.d              & o.o.d & i.i.d              & o.o.d    & i.i.d              & o.o.d             \\ \midrule

MDD \cite{lan2022closer} & TOMM'2023 & - & - & 52.78 & 40.39 & 34.71 & 22.70 & - & - & 43.63 & 20.80 & 31.44 & 11.66 \\ 
        Multi-NA \cite{lan2023curriculum} & AAAI'2023 & 64.21 & 52.21 & 53.82 & 39.86 & 34.47 & 21.38 & 49.91 & 32.32 & 41.67 & 20.78 & 28.82 & 11.03 \\ 
        DFM \cite{wang2023mixup} & ACM MM'2023 & 64.50 & 56.49 & 57.97 & 41.65 & 35.37 & 23.34 & 57.21 & 38.82 & 46.05 & 25.27 & 30.17 & 12.55 \\ 
        CDS \cite{qi2024collaborative} & TCSVT'2024 & - & - & 49.84 & 41.50 & 33.89 & 24.19 & - & - & 39.21 & 20.81 & 25.87 & 11.09 \\ 
        BSSARD* \cite{qi2024bias} & AAAI'2024 & 63.09 & 55.78 & 53.01 & 43.12 & 36.34 & 25.60 & 52.50 & 36.41 & 43.02 & 24.60 & 30.92 & 14.32 \\ 
        \midrule
        Baseline & ICASSP'2025 & 63.46 & 55.41 & 51.46 & 39.60 & 32.51 & 22.55 & 52.83 & 36.32 & 39.82 & 20.49 & 25.79 & 11.31 \\
        \rowcolor{blue!15} Ours & ICASSP'2025 & \textbf{68.65} & \textbf{58.76} & \textbf{58.74} & \textbf{44.61} & \textbf{37.75} & \textbf{26.81} & \textbf{58.01} & \textbf{39.52} & \textbf{46.91} & \textbf{25.89} & \textbf{31.54} & \textbf{14.44} \\ 
                          
         \bottomrule
\end{tabular}
\vspace{-1.5em}
\end{table*}

\section{Experiment}
\label{sec:Experiment}

\subsection{Experiment Setup}
\label{subsec:Experiment Setup}

\paragraph{Datasets}
Our experiments are conducted on Charades-CD and ActivityNet-CD, which are re-divided splits of Charades-STA \cite{gao2017tall} and ActivityNet Captions \cite{krishna2017dense} by \cite{lan2022closer}. The temporal distributions of samples in the training, val, and test-iid sets are independent and identically distributed (IID). Conversely, the test-ood set is specifically composed of out-of-distribution (OOD) samples to evaluate the generalization abilities of models across diverse temporal distributions.

\paragraph{Metrics} 
We adopt the commonly used R@$n$, IoU=$\theta$ as evaluation metrics. R@$n$, IoU=$\theta$ is the ratio of testing samples with at least one of the top-n localization results having an IoU score larger than $\theta$. We also report results with another metric dR@$n$, IoU=$\theta$ \cite{lan2022closer} which is discounted R@$n$, IoU=$\theta$ to restrain overlong predictions.

\paragraph{Implementation Details} We utilize 300d GloVe \cite{pennington2014glove} vectors to initialize word embeddings. The pre-trained I3D \cite{carreira2017quo} network is used to extract video features. The feature dimension $D$ is set to 256. $\beta_{sv}$ and $\beta_{lv}$ are both set to 10. The model is trained for 100 epochs using the Adam optimizer \cite{KingmaB14} with a learning rate of 0.0001 and batch size of 64. We set $\lambda_{\{1,2,3\}}$ to \{5, 1, 1\}.

\subsection{Comparison with State-of-the-Arts}

We compare our method with state-of-the-art methods after 2023 on Charades-CD and ActivityNet-CD in Table \ref{main_results}. 
Our method significantly improves the baseline’s grounding accuracy for all metrics in both the test-iid and test-ood sets. We also achieve the highest grounding accuracy for all metrics.

\subsection{Ablation Study}
\paragraph{Loss Terms} We study the effectiveness of each loss function and their combinations in Table \ref{Ablation_loss_terms}. The incorporation of either $L_{d}$ or $L_{kl}$ independently results in reductions on the baseline when data augmentation is applied. 
This is due to the model’s inability to discriminate the temporal discrepancy between the original and augmented videos. 
Moreover, their joint utilization effectively improves the performance by heightening the model’s awareness of temporal discrepancies between well-aligned original and augmented videos.

\begin{table}[t]
\scriptsize      
\centering
  \caption{Ablation study of loss terms on the test-ood set of Charades-CD. * denotes the absence of data augmentation.}
  \label{Ablation_loss_terms}
\begin{tabular}{cccccc}
\toprule
\multicolumn{3}{c}{Loss Terms} & \multicolumn{3}{c}{Charades-CD}  \\ 
\cmidrule(lr){1-3} \cmidrule(lr){4-6} 
                      $L_{loc}$      & $L_{d}$     & $L_{kl}$   & R@1,IoU=0.3       & R@1,IoU=0.5 & R@1,IoU=0.7      \\ \midrule
 \ \,\checkmark*      &            &     &   62.81 & 47.50    & 25.96          \\
 \checkmark      &            &       & 63.95 & 49.69    & 28.21         \\
 \checkmark      & \checkmark           &      &  63.17 & 48.63   & 25.99           \\
 \checkmark      &      & \checkmark            & 63.74 & 48.96   & 26.41           \\
 \rowcolor{blue!15}\checkmark      & \checkmark     & \checkmark     &    \textbf{65.60}      & \textbf{51.26}   & \textbf{29.68}            \\
\bottomrule
\end{tabular}
\end{table}

\paragraph{Data Augmentation Strategies} We investigate the contributions of each data augmentation strategy and their combinations in Table \ref{Ablation_Data_Augmentation_Strategies}. Each strategy achieves slightly better grounding performance upon the baseline model. The improvements are constrained since each strategy only enriches the temporal distribution from a specific perspective. Moreover, their joint application leads to further improvements by comprehensively diversifying the temporal distribution.

\begin{table}[t]
\scriptsize      
\centering
  \caption{Ablation study of data augmentation strategies on test-ood sets. \emph{SV} and \emph{LV} indicate shortened video augmentation and lengthened video augmentation, respectively.}
  \label{Ablation_Data_Augmentation_Strategies}
\begin{tabular}{cccccc}
\toprule
\multicolumn{2}{c}{Strategies} & \multicolumn{2}{c}{Charades-CD}     & \multicolumn{2}{c}{ActivityNet-CD}  \\
\cmidrule(lr){1-2} \cmidrule(lr){3-4} \cmidrule(lr){5-6}
                      SV             & LV             & R@1,IoU=0.5 & R@1,IoU=0.7  & R@1,IoU=0.5 & R@1,IoU=0.7   \\ \midrule
              &               & 47.50    & 25.96       & 25.81   & 13.81   \\
 \checkmark             &                & 48.42   & 26.14       & 25.94   & 13.87    \\
              & \checkmark               & 48.14   & 25.56     & 25.89   & 13.78    \\
\rowcolor{blue!15} \checkmark             & \checkmark                & \textbf{51.26}   & \textbf{29.68}        & \textbf{29.36}   & \textbf{16.25}    \\
\bottomrule
\end{tabular}
\vspace{-1.5em}
\end{table}

\begin{figure}[t]
\centering
\resizebox{0.9\linewidth}{!}{
  \subfloat{\hspace{-4mm} \includegraphics[width=0.50\linewidth]{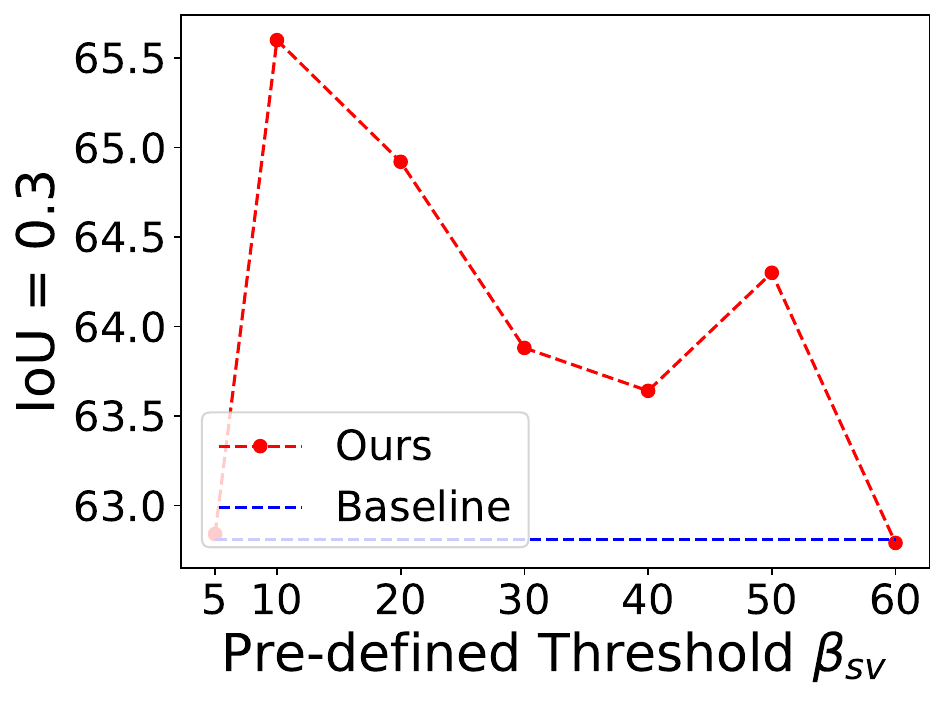}
  \hfill 
  \includegraphics[width=0.50\linewidth]{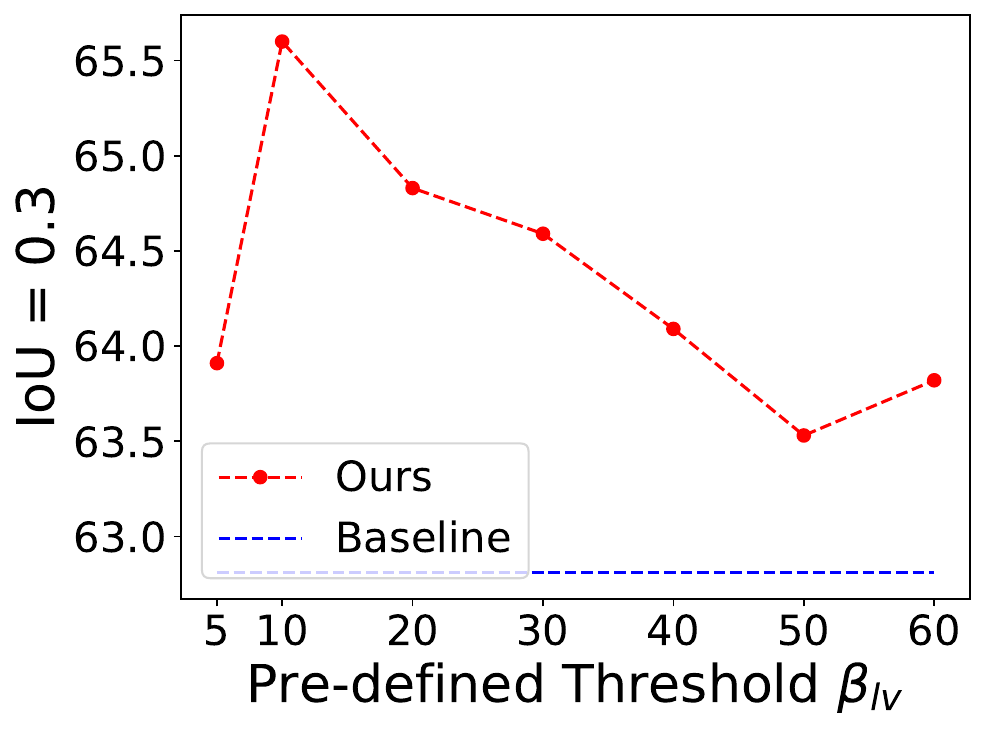}
  } 
}
\caption{Ablation studies of pre-defined thresholds $\beta_{sv}$ and $\beta_{lv}$ on the Charades-CD test-ood set with R@1, IoU=0.3.}
\label{fig:pre-defined thresholds}
\end{figure}

\paragraph{Pre-defined Thresholds} As depicted in Fig.  \ref{fig:pre-defined thresholds}, our method almost yields consistent enhancements in performance across the entire range of $\beta_{sv}$ and $\beta_{lv}$ values. Notably, as $\beta_{sv}$ or $\beta_{lv}$ increase, there is an initial uptick in performance, and overall, a general decline is observed. Our analysis suggests that minor adjustments to video lengths do not substantially affect the temporal distribution. Conversely, extensive modifications to longer clips can disrupt long-term temporal contexts and introduce noise into the training process. Optimal results are achieved when $\beta_{sv}$ and $\beta_{lv}$ are set between 5 and 20.

\paragraph{Grounding Structures} We verify the effectiveness of our training framework on backbones with different grounding structures.
As depicted in Table \ref{grounding_structures}, our method effectively improves the performance regardless of the structure, including proposal-free \cite{hao2022query}, proposal-based \cite{zhang2020learning} and DETR-based \cite{lei2021detecting} methods.
This proves that our training framework is model-agnostic, highlighting its generalization abilities.

\begin{table}[t!]
\scriptsize      
\centering
\caption{Effect on multiple grounding structures in test-ood sets with R@1, IoU=0.5. * indicates our reproduced results.}
\label{grounding_structures}
\begin{tabular}{lccccccccccccc}
\toprule
\multirow{1}{*}{Method}    & \multicolumn{1}{c}{Charades-CD}                                                           & \multicolumn{1}{c}{ActivityNet-CD} \\              

\midrule
QAVE \cite{hao2022query}   & 38.22    & 21.39  \\ 
\rowcolor{blue!15}+ Ours   & \textbf{47.58}   & \textbf{25.74}   \\
\midrule
2D-TAN \cite{zhang2020learning}   & 35.88    & 22.01   \\ 
\rowcolor{blue!15}+ Ours   & \textbf{39.91}   & \textbf{26.42}   \\
\midrule
Moment-DETR* \cite{lei2021detecting}   & 46.73    & 24.37   \\ 
\rowcolor{blue!15}+ Ours   & \textbf{54.89}    & \textbf{30.15}  \\ 
\bottomrule
\end{tabular}
\vspace{-1.5em}
\end{table}

\begin{figure}[h!]
\centering
  \subfloat{\includegraphics[width=0.5\linewidth]{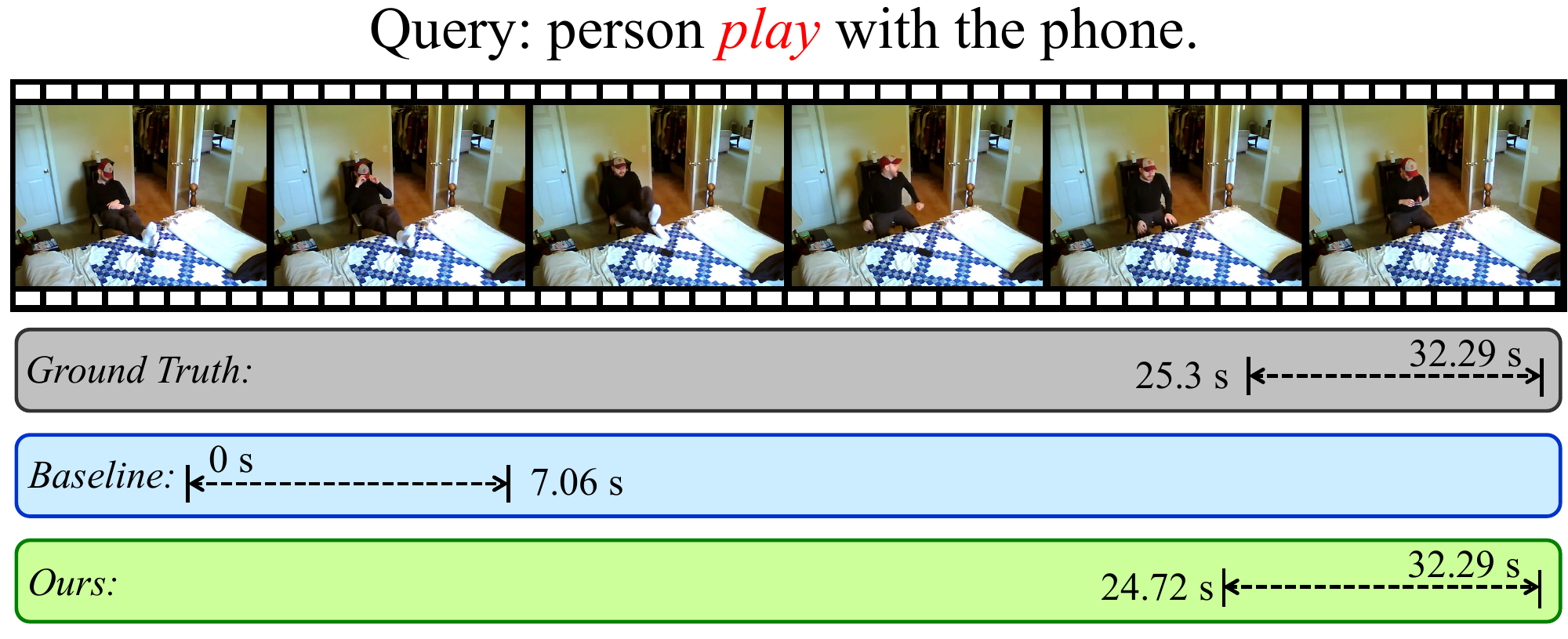}} 
  \hfill
  \subfloat{\includegraphics[width=0.205\linewidth]{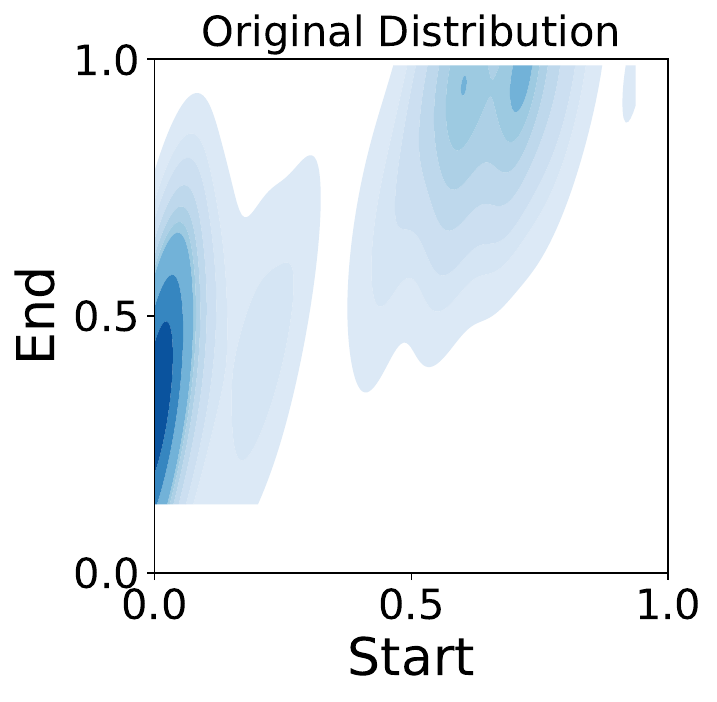}
  \includegraphics[width=0.272\linewidth]{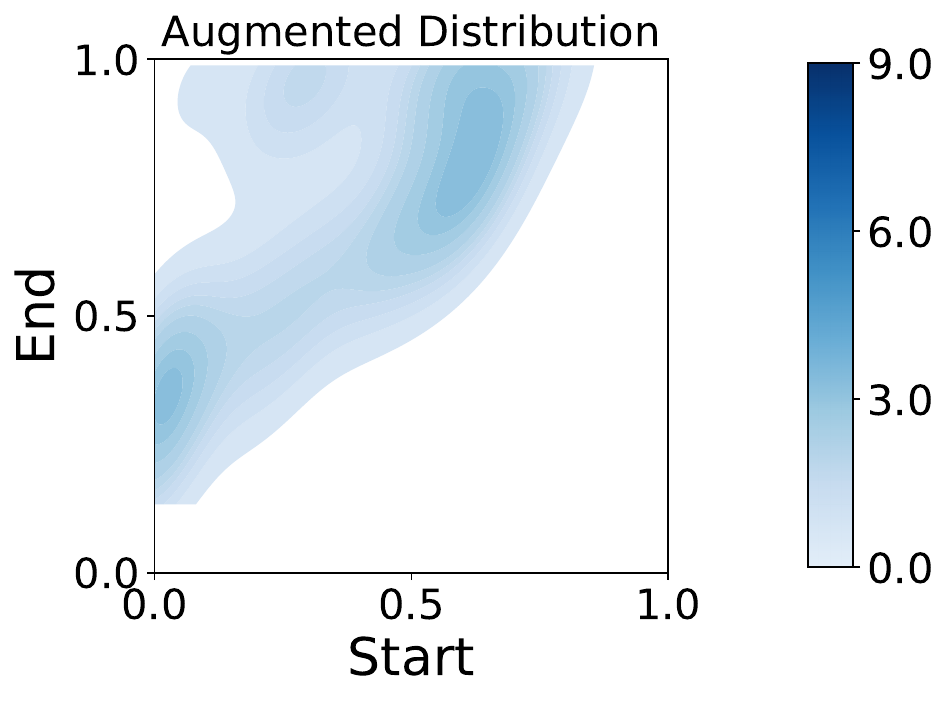}} 
   
\caption{The visualization comparison results between the baseline model and our method on Charades-CD.}
\label{Qualitative_Results}
\vspace{-1.5em}
\end{figure}

\subsection{Qualitative Results} We report an illustrative example of temporal bias on Charades-CD. As demonstrated in Fig. \ref{Qualitative_Results}, the majority of temporal locations for queries containing the verb \emph{play} tend to be the initial half of the videos. For a query containing the verb \emph{play}, if a model excessively depends on temporal biases in the training set for prediction, it is likely to predict the target moment in the first half of the video, despite the actual target moment residing in the latter segment. Concretely, the baseline model exhibits this tendency. In contrast, our method substantially diversifies the temporal distribution, leading to the accurate identification of temporal locations.

\section{Conclusion}
In this paper, we propose a novel debiased training framework for TSGV which is to generate and leverage videos with various lengths and target moment locations. It effectively diversifies temporal distributions of datasets, which in turn mitigates the model’s reliance on temporal biases inherent in datasets. We also propose a domain adaptation auxiliary task to mitigate the noise during training and improve the grounding precision. Extensive experiments on Charades-CD and ActivityNet-CD substantiate the effectiveness and robustness of our method in enhancing multiple grounding structures’ generalization capabilities, achieving state-of-the-art results.

\bibliographystyle{IEEEtran}
\bibliography{IEEEabrv,refs}

\begin{thebibliography}{10}
\providecommand{\url}[1]{#1}
\csname url@samestyle\endcsname
\providecommand{\newblock}{\relax}
\providecommand{\bibinfo}[2]{#2}
\providecommand{\BIBentrySTDinterwordspacing}{\spaceskip=0pt\relax}
\providecommand{\BIBentryALTinterwordstretchfactor}{4}
\providecommand{\BIBentryALTinterwordspacing}{\spaceskip=\fontdimen2\font plus
\BIBentryALTinterwordstretchfactor\fontdimen3\font minus \fontdimen4\font\relax}
\providecommand{\BIBforeignlanguage}[2]{{%
\expandafter\ifx\csname l@#1\endcsname\relax
\typeout{** WARNING: IEEEtran.bst: No hyphenation pattern has been}%
\typeout{** loaded for the language `#1'. Using the pattern for}%
\typeout{** the default language instead.}%
\else
\language=\csname l@#1\endcsname
\fi
#2}}
\providecommand{\BIBdecl}{\relax}
\BIBdecl

\bibitem{yuan2019semantic}
Y.~Yuan, L.~Ma, J.~Wang, W.~Liu, and W.~Zhu, ``Semantic conditioned dynamic modulation for temporal sentence grounding in videos,'' \emph{Advances in Neural Information Processing Systems}, vol.~32, 2019.

\bibitem{ghosh2019excl}
S.~Ghosh, A.~Agarwal, Z.~Parekh, and A.~Hauptmann, ``Excl: Extractive clip localization using natural language descriptions,'' in \emph{Proceedings of NAACL-HLT}, 2019, pp. 1984--1990.

\bibitem{zeng2020dense}
R.~Zeng, H.~Xu, W.~Huang, P.~Chen, M.~Tan, and C.~Gan, ``Dense regression network for video grounding,'' in \emph{Proceedings of the IEEE/CVF Conference on Computer Vision and Pattern Recognition}, 2020, pp. 10\,287--10\,296.

\bibitem{mun2020local}
J.~Mun, M.~Cho, and B.~Han, ``Local-global video-text interactions for temporal grounding,'' in \emph{Proceedings of the IEEE/CVF Conference on Computer Vision and Pattern Recognition}, 2020, pp. 10\,810--10\,819.

\bibitem{hu2021coarse}
Y.~Hu, L.~Nie, M.~Liu, K.~Wang, Y.~Wang, and X.-S. Hua, ``Coarse-to-fine semantic alignment for cross-modal moment localization,'' \emph{IEEE Transactions on Image Processing}, vol.~30, pp. 5933--5943, 2021.

\bibitem{wang2021structured}
H.~Wang, Z.-J. Zha, L.~Li, D.~Liu, and J.~Luo, ``Structured multi-level interaction network for video moment localization via language query,'' in \emph{Proceedings of the IEEE/CVF Conference on Computer Vision and Pattern Recognition}, 2021, pp. 7026--7035.

\bibitem{zhang2020span}
H.~Zhang, A.~Sun, W.~Jing, and J.~T. Zhou, ``Span-based localizing network for natural language video localization,'' in \emph{Proceedings of the 58th Annual Meeting of the Association for Computational Linguistics}, 2020, pp. 6543--6554.

\bibitem{wang2022negative}
Z.~Wang, L.~Wang, T.~Wu, T.~Li, and G.~Wu, ``Negative sample matters: A renaissance of metric learning for temporal grounding,'' in \emph{Proceedings of the AAAI Conference on Artificial Intelligence}, vol.~36, no.~3, 2022, pp. 2613--2623.

\bibitem{wang2023scene}
Z.~Wang, Y.~Zhao, H.~Huang, Y.~Xia, and Z.~Zhao, ``Scene-robust natural language video localization via learning domain-invariant representations,'' in \emph{Findings of the Association for Computational Linguistics: ACL 2023}, 2023, pp. 144--160.

\bibitem{lin2023univtg}
K.~Q. Lin, P.~Zhang, J.~Chen, S.~Pramanick, D.~Gao, A.~J. Wang, R.~Yan, and M.~Z. Shou, ``Univtg: Towards unified video-language temporal grounding,'' in \emph{Proceedings of the IEEE/CVF International Conference on Computer Vision}, 2023, pp. 2794--2804.

\bibitem{jiang2024prior}
Y.~Jiang, W.~Zhang, X.~Zhang, X.~Wei, C.~W. Chen, and Q.~Li, ``Prior knowledge integration via llm encoding and pseudo event regulation for video moment retrieval,'' in \emph{ACM Multimedia 2024}, 2024.

\bibitem{fang2024not}
X.~Fang, W.~Fang, D.~Liu, X.~Qu, J.~Dong, P.~Zhou, R.~Li, Z.~Xu, L.~Chen, P.~Zheng \emph{et~al.}, ``Not all inputs are valid: Towards open-set video moment retrieval using language,'' in \emph{ACM Multimedia 2024}, 2024.

\bibitem{otani2020challengesmr}
E.~R. Mayu~Otani, Yuta~Nakahima and J.~Heikkil{\"{a}}, ``Uncovering hidden challenges in query-based video moment retrieval,'' in \emph{The British Machine Vision Conference (BMVC)}, 2020.

\bibitem{lan2022closer}
X.~Lan, Y.~Yuan, X.~Wang, L.~Chen, Z.~Wang, L.~Ma, and W.~Zhu, ``A closer look at debiased temporal sentence grounding in videos: Dataset, metric, and approach,'' \emph{ACM Trans. Multimedia Comput. Commun. Appl.}, vol.~19, no.~6, jul 2023.

\bibitem{krishna2017dense}
R.~Krishna, K.~Hata, F.~Ren, L.~Fei-Fei, and J.~Carlos~Niebles, ``Dense-captioning events in videos,'' in \emph{Proceedings of the IEEE international conference on computer vision}, 2017, pp. 706--715.

\bibitem{hao2022can}
J.~Hao, H.~Sun, P.~Ren, J.~Wang, Q.~Qi, and J.~Liao, ``Can shuffling video benefit temporal bias problem: A novel training framework for temporal grounding,'' in \emph{European Conference on Computer Vision}.\hskip 1em plus 0.5em minus 0.4em\relax Springer, 2022, pp. 130--147.

\bibitem{lan2023curriculum}
X.~Lan, Y.~Yuan, H.~Chen, X.~Wang, Z.~Jie, L.~Ma, Z.~Wang, and W.~Zhu, ``Curriculum multi-negative augmentation for debiased video grounding,'' in \emph{Proceedings of the AAAI Conference on Artificial Intelligence}, vol.~37, no.~1, 2023, pp. 1213--1221.

\bibitem{qi2024bias}
Z.~Qi, Y.~Yuan, X.~Ruan, S.~Wang, W.~Zhang, and Q.~Huang, ``Bias-conflict sample synthesis and adversarial removal debias strategy for temporal sentence grounding in video,'' in \emph{AAAI}, 2024.

\bibitem{liu2022reducing}
D.~Liu, X.~Qu, and W.~Hu, ``Reducing the vision and language bias for temporal sentence grounding,'' in \emph{Proceedings of the 30th ACM International Conference on Multimedia}, 2022, pp. 4092--4101.

\bibitem{wang2023mixup}
X.~Wang, Z.~Wu, H.~Chen, X.~Lan, and W.~Zhu, ``Mixup-augmented temporally debiased video grounding with content-location disentanglement,'' in \emph{Proceedings of the 31st ACM International Conference on Multimedia}, 2023, pp. 4450--4459.

\bibitem{zhang2021towards}
H.~Zhang, A.~Sun, W.~Jing, and J.~T. Zhou, ``Towards debiasing temporal sentence grounding in video,'' \emph{arXiv preprint arXiv:2111.04321}, 2021.

\bibitem{qi2024collaborative}
Z.~Qi, Y.~Yuan, X.~Ruan, S.~Wang, W.~Zhang, and Q.~Huang, ``Collaborative debias strategy for temporal sentence grounding in video,'' \emph{IEEE Transactions on Circuits and Systems for Video Technology}, 2024.

\bibitem{vaswani2017attention}
A.~Vaswani, N.~Shazeer, N.~Parmar, J.~Uszkoreit, L.~Jones, A.~N. Gomez, {\L}.~Kaiser, and I.~Polosukhin, ``Attention is all you need,'' \emph{Advances in neural information processing systems}, vol.~30, 2017.

\bibitem{carreira2017quo}
J.~Carreira and A.~Zisserman, ``Quo vadis, action recognition? a new model and the kinetics dataset,'' in \emph{proceedings of the IEEE Conference on Computer Vision and Pattern Recognition}, 2017, pp. 6299--6308.

\bibitem{pennington2014glove}
J.~Pennington, R.~Socher, and C.~D. Manning, ``Glove: Global vectors for word representation,'' in \emph{Proceedings of the 2014 conference on empirical methods in natural language processing (EMNLP)}, 2014, pp. 1532--1543.

\bibitem{yu2018qanet}
A.~W. Yu, D.~Dohan, M.-T. Luong, R.~Zhao, K.~Chen, M.~Norouzi, and Q.~V. Le, ``Qanet: Combining local convolution with global self-attention for reading comprehension,'' in \emph{International Conference on Learning Representations}, 2018.

\bibitem{devlin-etal-2019-bert}
J.~Devlin, M.-W. Chang, K.~Lee, and K.~Toutanova, ``{BERT}: Pre-training of deep bidirectional transformers for language understanding,'' in \emph{Proceedings of the 2019 Conference of the North {A}merican Chapter of the Association for Computational Linguistics: Human Language Technologies, Volume 1 (Long and Short Papers)}.\hskip 1em plus 0.5em minus 0.4em\relax Minneapolis, Minnesota: Association for Computational Linguistics, Jun. 2019, pp. 4171--4186.

\bibitem{ganin2016domain}
Y.~Ganin, E.~Ustinova, H.~Ajakan, P.~Germain, H.~Larochelle, F.~Laviolette, M.~March, and V.~Lempitsky, ``Domain-adversarial training of neural networks,'' \emph{Journal of machine learning research}, vol.~17, no.~59, pp. 1--35, 2016.

\bibitem{gao2017tall}
J.~Gao, C.~Sun, Z.~Yang, and R.~Nevatia, ``Tall: Temporal activity localization via language query,'' in \emph{Proceedings of the IEEE international conference on computer vision}, 2017, pp. 5267--5275.

\bibitem{KingmaB14}
D.~P. Kingma and J.~Ba, ``Adam: {A} method for stochastic optimization,'' in \emph{3rd International Conference on Learning Representations}, 2015.

\bibitem{hao2022query}
J.~Hao, H.~Sun, P.~Ren, J.~Wang, Q.~Qi, and J.~Liao, ``Query-aware video encoder for video moment retrieval,'' \emph{Neurocomputing}, vol. 483, pp. 72--86, 2022.

\bibitem{zhang2020learning}
S.~Zhang, H.~Peng, J.~Fu, and J.~Luo, ``Learning 2d temporal adjacent networks for moment localization with natural language,'' in \emph{Proceedings of the AAAI Conference on Artificial Intelligence}, vol.~34, no.~07, 2020, pp. 12\,870--12\,877.

\bibitem{lei2021detecting}
J.~Lei, T.~L. Berg, and M.~Bansal, ``Detecting moments and highlights in videos via natural language queries,'' \emph{Advances in Neural Information Processing Systems}, vol.~34, pp. 11\,846--11\,858, 2021.

\end{thebibliography}

\end{document}